\documentclass[final]{cvpr}

\usepackage{times}
\usepackage{epsfig}
\usepackage{graphicx}
\usepackage{amsmath}
\usepackage{amssymb}
\usepackage{ifthen}
\usepackage{bbm}

\usepackage[pagebackref=true,breaklinks=true,colorlinks,bookmarks=false]{hyperref}

\def\cvprPaperID{4209} 
\def\confYear{CVPR 2021}

\usepackage{array}
\newcolumntype{P}[1]{>{\centering\arraybackslash}p{#1}}
\DeclareMathOperator*{\argmax}{\arg\!\max}

\begin{document}

\newboolean{showcomments}
\setboolean{showcomments}{true}
\newcommand{\ka}[1]{\ifthenelse{\boolean{showcomments}} {\textcolor{magenta}{(Ashutosh says: #1)}} {} }
\newcommand{\sk}[1]{\ifthenelse{\boolean{showcomments}} {\textcolor{red}{(Saurabh says: #1)}} {} }

\newcommand{\blue}[1]{\textcolor{blue}{#1}} 
\newcommand{\green}[1]{\textcolor{green}{#1}}

\title{3D-NVS: A 3D Supervision Approach for Next View Selection}

\author{Kumar Ashutosh\\
IIT Bombay\\
{\tt\small kumar.ashutosh@iitb.ac.in}
\and
Saurabh Kumar\\
IIT Bombay\\
{\tt\small saurabhkm@iitb.ac.in}

\and
Subhasis Chaudhuri\\
IIT Bombay\\
{\tt\small sc@ee.iitb.ac.in}
}

\maketitle


\begin{abstract}

We present a classification based approach for the next best view selection and show how we can plausibly obtain a supervisory signal for this task. The proposed approach is end-to-end trainable and aims to get the best possible 3D reconstruction quality with a pair of passively acquired 2D views. The proposed model consists of two stages: a classifier and a reconstructor network trained jointly via the indirect 3D supervision from ground truth voxels. While testing, the proposed method assumes no prior knowledge of the underlying 3D shape for selecting the next best view. We demonstrate the proposed method's effectiveness via detailed experiments on synthetic and real images and show how it provides improved reconstruction quality than the existing state of the art 3D reconstruction and the next best view prediction techniques.
\end{abstract}

\section{Introduction}
Three dimensional (3D) object reconstruction and acquisition are a long time focus of machine vision research and a key to visual understanding and interpretation. It has a wide range of practical applications, including robotics, AR/VR, autonomous navigation, and industrial automation, to name just a few.
Being able to access the 3D geometric cue about an object or scene can immensely benefit in post-processing and decision making for instance in recognition \cite{pontil1998support}, segmentation \cite{qi2017pointnet}, pose estimation \cite{andriluka2010monocular} and other computer vision and pattern recognition tasks.

As a computer vision task, this requires a sensor for observing the object, which can be of two types, namely, active and passive cameras.
Active cameras like Laser imaging detection and ranging (LiDAR), Time of Flight (ToF), and depth cameras like the Kinect directly acquire scene depth and, in turn, the 3D shape of an object but come with their limitations apart from the high cost and power requirements.
Passive cameras are much cheaper and use significantly less power to acquire two-dimensional (2D) optical RGB snapshots of the object, which can then be used to recover the 3D structure by posing this as an inverse problem.

Recovering 3D shape from its two-dimensional snapshots has been a goal of classic computer vision problems like multiview stereo and Shape-from-X techniques.
Early methods approached this with a geometric perspective and focused on mathematically modeling the process of 2D projection to pose it as an ill-posed problem \cite{hartley2003multiple} constrained by suitable scene priors.
The structure from motion(SfM) techniques can be used to obtain a sparse 3D point cloud of an object along with the respective camera poses.
Along similar lines, Multi-View Stereo (MVS) techniques can provide a dense 3D reconstruction.
Recently, with the advent of deep learning-based methods, a model can directly learn a non-linear mapping from the images to the 3D representation outperforming classical methods while doing it in real-time \cite{pix2vox}.

Although there are works that attempt to reconstruct the 3D shape of an object from just a single view but these are in early stages due to their highly ill-posed nature of the problem.
Moreover, as in MVS, taking more than one view of the object reduces ambiguity and occlusion to significantly improve the 3D recognition quality as more information is available for recovery.
Typically, these views are acquired from a set of passively selected viewpoints independent of the object geometry.
To obtain the best 3D reconstruction, one would expect these viewpoints to depend on the 3D shape in consideration.
Therefore, to obtain the best possible 3D reconstruction quality from the captured images, one must actively decide the next best viewpoint to take the snapshot from.

The Next View Selection (NVS) problem has been explored since the early days of machine vision research \cite{connolly1985determination}.
This problem arises in many scenarios where one would want to find the next best viewpoint around an object or in an arena while maximizing a particular quality or performance metric.
For instance, to better grasp an object, a robotic arm equipped with a camera may want to find the next viewpoint to obtain the best 3D reconstruction for better understanding quickly.
Similarly, a UAV hovering over a scene may want to find the next view, which can provide the best 3D digital elevation model to minimize the flight time.
Several works in robotic navigation, robotic arm positioning, and remote sensing are discussed in the related work section.

However, almost all of these works use range cameras to capture scene depth directly and select the next view either heuristically or by optimizing for the surface area covered, or information gained.
Along with this, they have aimed to reconstruct a surface representation of the scene like mesh or a point cloud.
Volumetric representation like voxels is better suited for practical applications due to ease of manipulation and training than surface or mesh-based representations.
Hence, we focus on learning a volumetric representation of the scene in this work.
Moreover, selecting the next views using only passive camera images of a scene has been explored much recently but not very actively in the literature. We focus on this particular setup in this paper.
Predicting a 3D shape or next views using passive cameras is a more challenging task due to much lesser data available for use than in active sensors.
Due to this setup, classical solutions based on the surface reconstruction and optimized for the surface area covered or information gained cannot be used here.

This paper attempts to address these gaps and propose a novel classification-based approach for NVS without using the preprocessed ground truth next views.
Since our goal is to maximize the 3D reconstruction quality, we propose to extract a supervisory signal from the reconstruction process itself to drive the classification model's training and obtain a better loss-goal correspondence.
We present how a light-weight neural network-based classifier can be incorporated into existing 3D reconstruction models to perform NVS and trained without the knowledge of ground truth NVS labels.
Training such a classifier-reconstructor architecture is not straightforward without the labels. Therefore, we propose a novel loss objective that facilitates end to end training of such a joint classifier-reconstructor deep neural network model.

We demonstrate how the proposed method successfully selects the next views to maximize the reconstruction quality for a given 3D object on synthetic and real data.
Additionally, we analyze the selected next views for each object category to gain insights and observe underlying patterns to learn how they are distributed depending on the object shape.
Our contributions in this paper are as follows:
\begin{itemize}
    \item We propose a deep learning-based approach for next view selection and show how it can be effectively approached as a classification problem.
    \item We show how the supervisory signal from the 3D reconstruction process can be used to learn both the classification and reconstruction networks jointly in an end to end manner.
    \item We present a qualitative analysis of the selected next views to gain insights into the process and their dependence on object categories and shape.
    \item We present exhaustive experiments on both synthetic and real data to demonstrate the effectiveness of the proposed method in selecting appropriate next views while providing high reconstruction quality.
\end{itemize}

\begin{figure*}
\begin{center}
\includegraphics[scale=0.36]{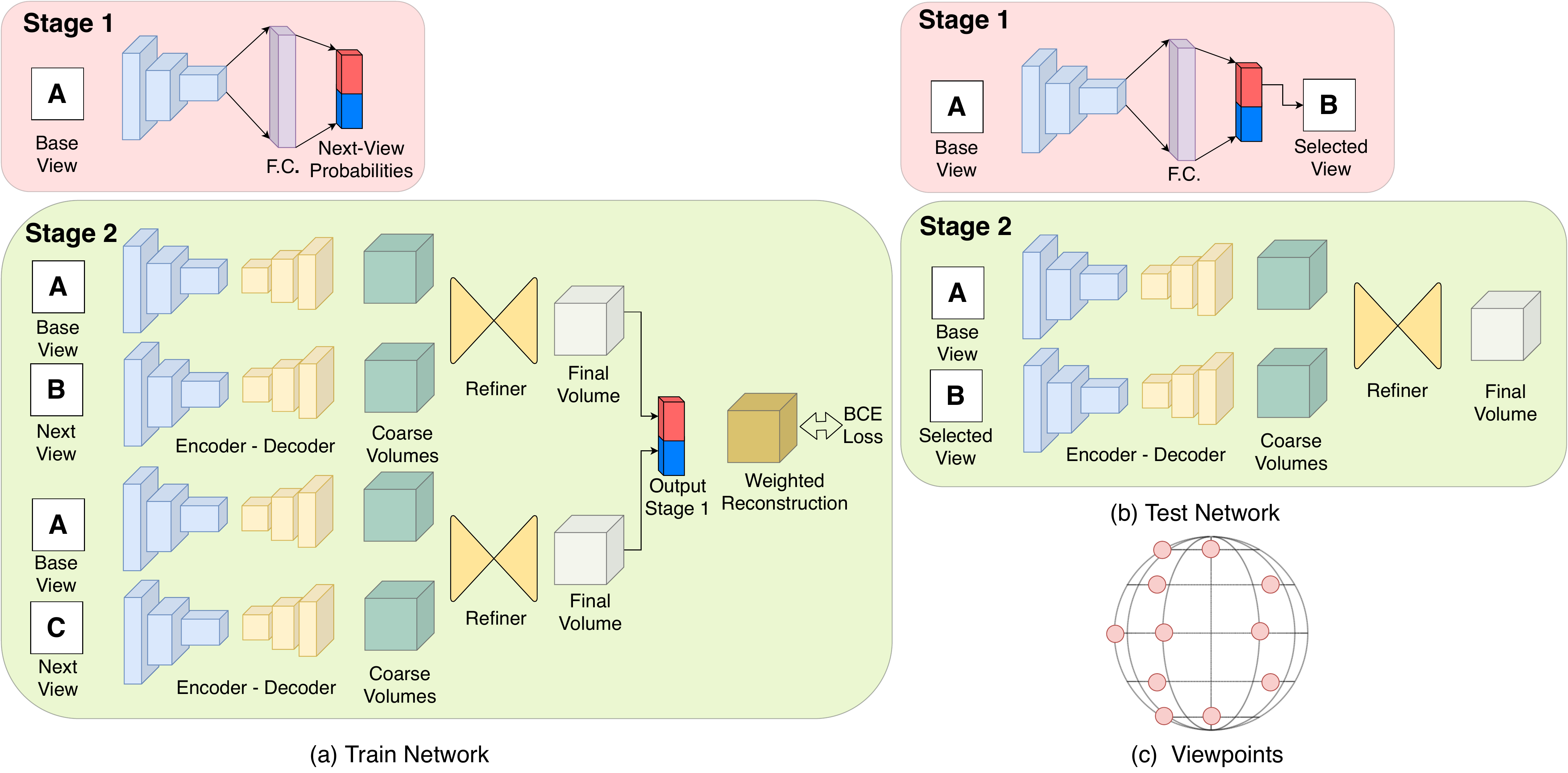}
\end{center}
   \caption{An Overview of the proposed 3D-NVS model used during the (a) training and (b) testing. A detailed explanation of each sub-parts is given in Section \ref{sec:method}. (c) Chosen viewpoints of the ShapeNet dataset to generate images for training.}
\label{fig:overview-model}
\end{figure*}

\section{Related Work}
Early works on 3D reconstruction focus on mathematically modeling the imaging process and solving an ill-posed problem via suitable scene priors and geometric constraints \cite{hartley2003multiple}.
Among the recent deep learning-based methods, there are three broad types of techniques based on output representation: Volume-based, Surface-based, and Intermediate representation.
Surface-based methods explore meshes and point clouds that are unordered representations \cite{wang2018pixel2mesh, kanazawa2018learning, pontes2018image2mesh}. 
Volumetric methods explore regular voxel grid-based representation as we consider in this paper \cite{wu2017marrnet, tulsiani2017multi, yang2018learning}.
The recent state of the art works in this category include Pix3D \cite{sun2018pix3d}, 3DR2N2 \cite{3dr2n2} and Pix2Vox \cite{pix2vox} which we use as our baselines to compare with.
Some techniques decompose the problem into sequential steps, each predicting a different representation that can be used as per application \cite{tatarchenko2016multi, lin2018learning}.


Early works in NVS acquire range images and aim to recover the surface geometry of an object.
Connolly et al. \cite{connolly1985determination} did one of the first works on automatic range sensor positioning using partial octree models and optimize for the surface area covered.
Maver et al. employ occlusion guidance \cite{maver1993occlusions} and optimize for information gained \cite{maver1993planning} for NVS.
These and related methods \cite{pito1996sensor, pito1995solution, pito1999solution} require the images to overlap to allow for registration and integration of the new data captured with previous scans. 
The earliest work in data-driven NVS is by Banta et al. \cite{banta1995best}, which uses synthetic range data to predict occupancy grids.
Here, the number of viewpoints on the view sphere is limited, and the radius is large enough to encompass whole object.
The center and coordinate axes of the view sphere are aligned with those of the object, and the camera always faces this center.
We work with a similar setup but use passive RGB cameras.

Recent works also focus on active sensors such as Hepp et al. \cite{hepp2018learn} use a CNN to learn a viewpoint utility function, Delmarico et al. \cite{delmerico2018comparison} optimize volumetric information gain, and Mendez et al. \cite{mendez2017taking} maximize information gain in terms of total map coverage and reconstruction quality.
Chen et al. \cite{chen2005vision} propose a prediction model for the object's unknown area using a 3D range data acquisition.
Zhang et al. \cite{delmerico2018comparison} use self-occlusion information for NVS to predict 3D motion estimation.
Potthast et al. \cite{potthast2014probabilistic} and Isler et al. \cite{isler2016information} propose information gain based NVS for occluded enviroments and 3D reconstruction respectively.
Other works include greedy \cite{obwald2018gpu} and supervised learning approach \cite{ly2019autonomous} for visibility based exploration and reconstruction of 3D environments.
Hashim et al. \cite{hashim2019next} propose a view selection recommendation method by minimizing pose ambiguity for object pose estimation using depth sensors.
There are also works on robotic arm navigation for autonomous exploration and grasping \cite{monica2018contour, morrison2019multi, bai2017toward, wu2019plant}, UAV positioning in remote sensing \cite{almadhoun2019guided}, Pose estimation \cite{sock2017multi} and 3D scene inpainting \cite{han2019deep}.

\begin{figure*}
\begin{center}
\includegraphics[scale=0.43]{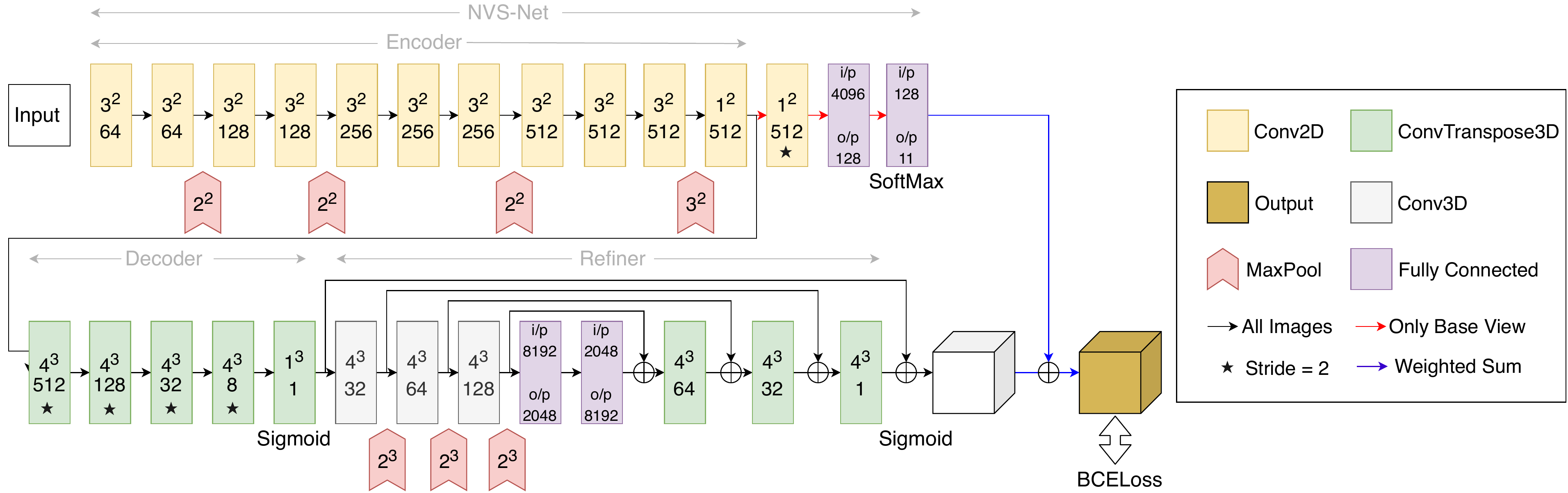}
\end{center}
   \caption{The proposed 3D-NVS architecture showing various constituent stages: NVS-Net, Encoder, Decoder and Refiner modules. Note that Encoder shares weight with NVS-Net in the 3D-NVS model for efficient parameter utilization leading to a smaller model size.}
\label{fig:model-architecture}
\end{figure*}

Closely related to our work and among the deep learning-based methods, Wu et al. \cite{wu20153d} propose a deep neural network-based view prediction but use 2.5D RGB-D images to improve 3D object recognition.
Daudelin et al. \cite{daudelin2017adaptable} propose a probabilistic NVS algorithm but use RGB-D sensor data for 3D object reconstruction.
Mendez et al. \cite{mendez2016next} use stereo image pairs and propose NVS to optimize the surface area covered by reconstructing the object surface.
Dunn et al. \cite{dunn2009next} use passive cameras but optimize for the surface area covered as their goal is surface reconstruction.
Mendoza et al. \cite{mendoza2020supervised} also pose this as a classification problem but work with range sensor data and generate exhaustive ground truth data to train a classifier.
They use a small dataset of 12 objects and a CNN-based classifier that takes the partial voxels as input to predict the next view. 
In contrast, we use RGB images and do not generate the ground truth but propose a classification-based approach guided by the supervision from 3D reconstruction directly.
Yang et al. \cite{yang2018active} propose a recurrent network for view prediction using passive cameras but use a highly restricted set of viewpoints on the view sphere, unlike the broad viewpoints and restrained acquisition process considered in the literature.
\section{Method}
\label{sec:method}
This section introduces 3D-NVS -- a novel deep learning model to reconstruct 3D shapes from 2D images while selecting the best next view. The proposed two-stage model works as follows: given an input image, stage 1 selects the next camera position out of 11 possible labels. Subsequently, stage 2 inputs the two images (base image and the selected next view) to a reconstruction model to generate a 3D shape. The shape is in voxel grids with binary occupancy (1 means occupied, 0 means empty). To select the next views, 3D-NVS uses supervision from ground truth 3D shapes. This indirect supervision is facilitated by a novel loss function, which enables the training of a network with a softmax classifier in between. Such a unified classifier followed by reconstruction architecture is the first in our knowledge and is a key enabler of our proposed method. An overview of the proposed model is presented in Figure \ref{fig:overview-model}, and the detailed architecture is shown in Figure \ref{fig:model-architecture}.
\subsection{Network Architecture}
We use slightly different architectures for training and testing. For training, we incorporate the availability of images from all camera viewpoints. However, during testing, we only input images from one viewpoint, and the model predicts the next view. This idea reinforces the freedom of obtaining multiple images from all viewpoints during training while restricted access during testing. 

During training, the input RGB image from an arbitrary camera position, called henceforth as the base image, is passed into NVS-Net, which outputs a probability distribution over the next views. The number of possible camera positions for the next view is fixed (details in Section \ref{sec:experiments}) and, thus, is a classification problem. In parallel to NVS-Net, we pass images from all the viewpoints into the Encoder -- a sub-network of Stage 2 of 3D-NVS that generate feature maps. These are then used by the Decoder to produces coarse voxels. The context-aware fusion and the refiner \cite{pix2vox} are used to fine-tune the generated voxels. The 3D volumes obtained from all pairs of the base image and possible next-best views are combined with the probability distribution predicted by the NVS-Net to get a final 3D volume.

During testing, we detach stage 1, which is NVS-Net, from the training setup \& use it for the next view selection described above and predicts the probability distribution over the next views. Now, the viewpoint with maximum probability is selected as the next view, and the two views (base view and selected next view) are passed into the Encoder, Decoder, Context-aware Fusion, and Refiner sub-networks to obtain the final reconstructed 3D volume.
\subsubsection{NVS-Net}
We use a classification network inspired by VGG16 \cite{vgg} and modify the last few layers of this network to suit our output classes. We take the model pre-trained on ImageNet \cite{imagenet} and freeze the first nine layers. An RGB square input image of size 228 pixels is progressively passed into convolutional layers with ELU activations while increasing the number of channels as described in Figure \ref{fig:model-architecture}. Finally, we obtain a 256x4x4 feature map which is flattened into a 4096-dimensional fully-connected layer and fed to two further fully-connected layers output 128 and 11 channel outputs, respectively. The last layer's activation is set to softmax to obtain probability distribution over the next views, as in other standard classification models.

In the remainder of the paper, we define 3D-NVS-R as the model where we choose the next view randomly from all the available views. Also, we define 3D-NVS-II when we choose the second best view, 3D-NVS-III for the third view, and so on. Finally, 3D-NVS-F is when the next view is chosen to maximize the distance between the input images.
\subsubsection{Encoder-Decoder}
The Encoder extracts feature from both input RGB images, the base, and the next view image selected by NVS-Net. Since the NVS-Net already extracts features and flattens them for classification, we share its parameters with Encoder for parameter efficiency and smaller model. The output of the Encoder is a feature map of size 256x8x8, which is an intermediate output of NVS-Net. Note that only the base image is passed into the NVS-Net, whereas all the images are passed into the Encoder in stage 2 during the training phase.

After obtaining the feature maps, we use those as input to recover 3D shapes. The architecture of having an Encoder-Decoder model is motivated by its success in Image Segmentation tasks \cite{7803544} and to more related 3D shape reconstruction tasks \cite{pix2vox}. The feature maps of size 256x8x8 are resized to 2048x2x2 and operated on successive transposed convolution layers that decrease the number of channels and increase the feature sizes. The obtained 32x32x32 vector is passed through a sigmoid activation function to obtain occupancy probability for each cell.
\subsubsection{Context-aware Fusion and Refiner}
Context-aware Fusion and Refiner is used in Pix2Vox \cite{pix2vox}  to convert coarse 3D volumes into an improved 3D volume. The idea behind such an additional network is to account for different reconstructions from different viewpoints. The Context-Aware Fusion generates a score map based on the viewpoint, and the final volume is obtained by multiplying a softmax-based activation with the corresponding coarse volume from that view. Finally, the Refiner is a UNet-like architecture \cite{unet}, which acts as a residual network to correct remaining mis-recovered parts of the object. The activation of the final layer is again set to sigmoid to obtain occupancy probability for the 32x32x32 grid.
\subsubsection{Loss Function}
We propose a loss function to train this unique architecture. The proposed loss function considers both NVS-Net output probability and the reconstructed 3D shape for each pair of views. Let the probability of selecting $i^{th}$ image is $p_i$ as per NVS-Net. Correspondingly, let $v_i$ denote the reconstructed volume of size 32x32x32 using base image and $i^{th}$ image as input. The averaged 3D shape is defined as 
\begin{equation}
    r = \sum_{\forall i} p_i v_i
\end{equation}
We propose the averaged per-voxel binary cross entropy between $r$ and ground truth voxel $t$ as the loss function of the network. Mathematically,
\begin{equation}
    L(r, t) = \frac{1}{N} \sum_{j=1}^{N} (t_j\log(r_j) + (1-t_j)\log(1-r_j))
\end{equation}
It is advantageous to have output from both the sub-networks in the expression of the loss. Such a choice prevents a network weight bottleneck at the NVS-Net outputs and serves the objective of simultaneous classification and reconstruction tasks. Our motivation to use this loss function is to improve the weighted reconstruction; that is, at each training step, we aim to improve the reconstruction performance on the network. Once it is trained successfully, the NVS-Net is inclined to assign a higher weight to the view which has a lower loss, thus achieving our two-fold objectives of Next View Selection and reconstruction.

\section{Experiments and Observations}
\label{sec:experiments}
We conduct experiments with the proposed model on both the synthetic dataset and real images. The details of the dataset, implementations, and observations are presented in the next sub-sections.

\begin{table*}
\begin{center}
{\renewcommand{\arraystretch}{1.1} 
\begin{tabular}{P{0.14\textwidth}P{0.12\textwidth}P{0.12\textwidth}P{0.24\textwidth}P{0.11\textwidth}P{0.11\textwidth}}
\hline
Category & 3D-R2N2 \cite{3dr2n2} & Pix2Vox \cite{pix2vox} & 3D-NVS (Proposed) & 3D-NVS-R & 3D-NVS-F\\
\hline
airplane (810) & 0.486 & 0.619 & \textbf{0.627} & 0.418 & 0.384\\
bench (364) & 0.336 & 0.442 & \textbf{0.522} & 0.345 & 0.311\\
cabinet (315) & 0.645 & \textbf{0.709} & 0.656 & 0.619 & 0.595\\
car (1501) & 0.781 & 0.815 & \textbf{0.836} & 0.740 & 0.730\\
chair (1357) & 0.410 & 0.473 & \textbf{0.526} & 0.362 & 0.306\\
display (220) & 0.433 & \textbf{0.508} & \textbf{0.508} & 0.416 & 0.402\\
lamp (465) & 0.272 & 0.365 & \textbf{0.370} & 0.318 & 0.299\\
speaker (325) & 0.619 & \textbf{0.649} & 0.631 & 0.585 & 0.583\\
rifle (475) & 0.508 & \textbf{0.597} & 0.579 & 0.387 & 0.344\\
sofa (635) & 0.568 & 0.602 & \textbf{0.640} & 0.503 & 0.456\\
table (1703) & 0.405 & 0.438 & \textbf{0.510} & 0.343 & 0.293\\
telephone (211) & 0.636 & \textbf{0.818} & 0.681 & 0.628 & 0.700\\
watercraft (389) & 0.477 & 0.529 & \textbf{0.553} & 0.428 & 0.387\\
\hline 
Overall (8770) & 0.511 & 0.574 & \textbf{0.601} & 0.465 & 0.432\\
\hline
\end{tabular}
}
\end{center}
\caption{Class-wise comparison between state of the art reconstruction methods and our proposed 3D-NVS. Details of the experiment are described in Section \ref{sec:synthetic-results}. We also compare 3D-NVS-R and 3D-NVS-F with 3D-NVS, details of which is included in the Ablation Study (see Section \ref{sec:ablation}). We can observe that 3D-NVS has the best overall reconstruction IoU amongst all the methods.}
\label{tab:result}
\end{table*}

\subsection{Dataset}
\label{sec:dataset}
ShapeNet \cite{shapenet} is a large-scale dataset containing 3D CAD models of common objects. We use a subset of ShapeNet with 13 categories having 43,782 models. This subset choice is the same as that used in \cite{3dr2n2, pix2vox, yang2018active}. The 3D models are voxelized into a 32x32x32 grid using the binvox package \cite{binvox, nooruddin03}. However, the rendered images for previous work \cite{pix2vox, 3dr2n2, yang2018active} positions the camera only on a circular trajectory over a fixed elevation (25$^\circ$ -- 30$^\circ$). Such highly-constrained acquisition of images is not particularly applicable to real-world acquisition followed by reconstruction. In fact, we show that if these previous models are trained and tested on images taken from positions away from such fixed locations, the reconstruction performance drops significantly. Besides, many real-world applications, such as UAV acquisition and underwater acquisition, can have images from arbitrary locations, and requiring them to obtain images from a particular viewpoint may be infeasible.

To mitigate this inconsistency and to increase the robustness of chosen camera positions, we render images from a much broader viewpoint. We use PyTorch3D \cite{pytorch3d} to render images from much broader viewpoints with Soft Phong Shader. We choose elevations of $-90^{\circ}, -60^{\circ}, -30^{\circ}, 0^{\circ}, 30^{\circ}, 60^{\circ}, 90^{\circ}$ and the azimuths also vary uniformly around the sphere to generate 24 views that span the whole view sphere to represent a real acquisition scenario. Also, some of the object textures are incompatible with PyTorch3D, and thus, we use those models without adding any default texture. This choice also adds robustness to both textured and untextured images. We finally use 11 views in this experiment since diametrically opposite images are degenerate for untextured images (images from both poles are also removed to maintain mathematical stability during rendering calculations). A schematic representation of the camera locations is provided in Figure \ref{fig:overview-model}.

To evaluate our proposed model's performance on real images, we use the Pix3D dataset \cite{pix3d}. Pix3D offers real-world images of common objects, along with a 3D model of the same. The images are accompanied by a mask for the respective object class. Multiple images are corresponding to one 3D object taken from various viewpoints. Although viewpoint-mapping is not provided explicitly, we use the input image metadata's camera position and map it to the nearest viewpoint used in the previous synthetic data. However, we observe that the dataset does not contain as many images to cover the whole viewpoint. In such a scenario, the next-best view is the one that has the highest probability out of the available viewpoints.

\subsection{Evaluation Metrics}
We use the standard Intersection over Union (IoU) as the evaluation metrics. As the name suggests, this metric calculates the ratio of the number of matching occupied voxels and the total number of occupied voxels. Mathematically,
\begin{equation}
   IoU = \frac{\sum\limits_{\forall x, \forall y, \forall z} \mathbbm{1}_{\{r_{(x,y,z)}>v_t\}} \land t_{(x,y,z)} }{ \sum\limits_{\forall x \forall y \forall z} \mathbbm{1}_{r_{(x,y,z)}>v_t} \lor t_{(x,y,z)} }
\end{equation}

Here $v_t$ denotes the binarization threshold and $\land$, $\lor$ denote logical AND and OR, respectively. Also, $r_{(x,y,z)}$ and $t_{(x,y,z)}$ denotes values at location $(x,y,z)$ for predicted voxel grid and ground truth, respectively. IoU ranges from 0 to 1, with higher value implying better reconstruction.

\subsection{Network Parameters}
We use PyTorch \cite{pytorch_paper} to train 3D-NVS described in the previous section. The network is trained until convergence, which takes about 30 hours on an Nvidia 1080Ti GPU. The number of epochs is set to 20 with a batch size of 8. The learning time per epoch is high due to an increased number of gradient calculations for one sample. For all the possible next views, the model parameters are shared, and hence the network computes gradients for all the images. Fortunately, it also leads to a considerable increase in learning per epoch. The learning rate of the network layers is set to 0.0001. We experiment with both SGD optimizer, and Adam optimizer \cite{kingma:adam} and report the latter due to better performance. We use standard data augmentation techniques like adding random jitter and using random colored background for better robustness. To recall, the input images are of size 228x228, and the output shape is a voxel grid of size 32x32x32.

\begin{figure*}
\begin{center}
\includegraphics[scale=0.22]{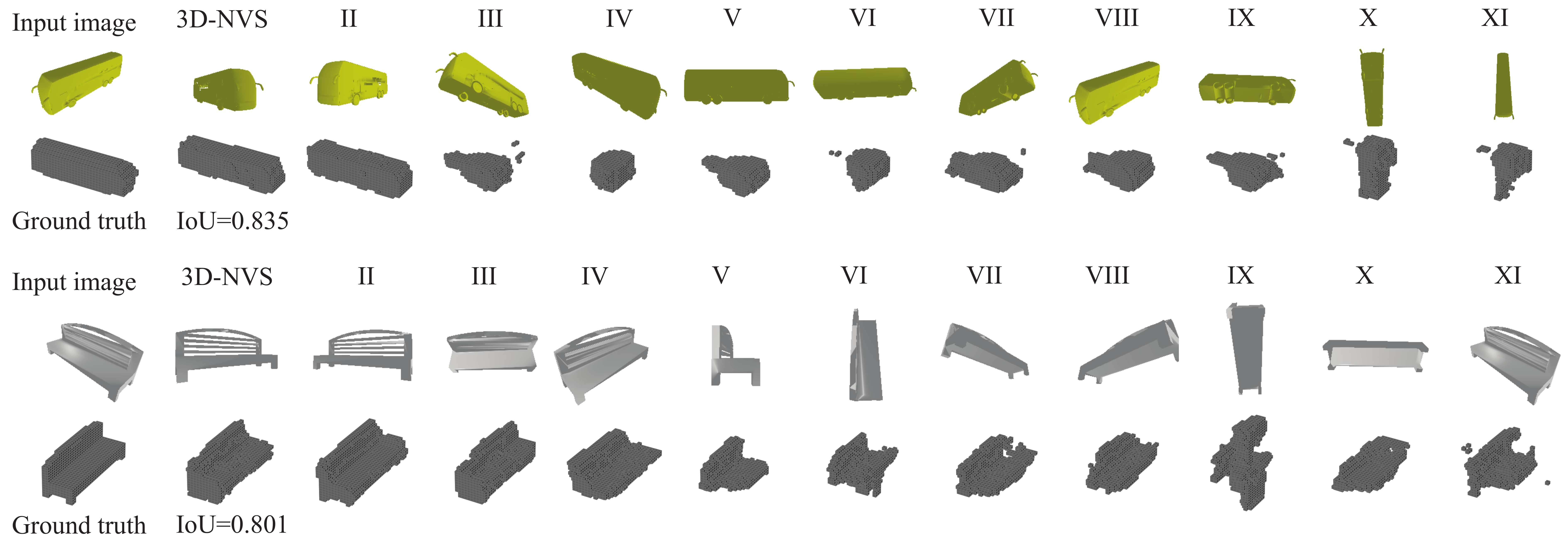}
\end{center}
   \caption{Examples of rendered input images and corresponding reconstruction performance. The images are sorted according to the probability outputs of the proposed NVS-Net. The voxel grids below them represent the reconstruction performance with that view along with the input image. We see a good reconstruction for high probability views, and the performance decreases as we choose less probable next views, showing that the selected next view is indeed the best. (Please zoom-in to compare voxel grids).}
\label{fig:sample_ouputs}
\end{figure*}

\subsection{Reconstruction Evaluation}
\label{sec:synthetic-results}
\subsubsection{Synthetic Multi-view Images}
To establish the effectiveness of our approach, we compare the reconstruction performance on various state of the art shape reconstruction methods. We conduct the experiment on rendered images discussed in Section \ref{sec:dataset}. Also, we again train 3D-NVS on the limited-elevation rendered images used in Pix2Vox \cite{pix2vox}, 3D-R2N2 \cite{3dr2n2}, and Yang et al. \cite{yang2018active} for a fair baseline and to demonstrate the robustness of our model with respect to capturing locations.

A comparison of our results with that of 3D-R2N2 and Pix2Vox is included in Table \ref{tab:result}. All the methods are trained on our rendered images with wider viewing angles and a mix of textured and untextured images. The proposed 3D-NVS  achieves a mean IoU of 0.601, thus outperforming both Pix2Vox and 3D-R2N2, which achieves a mean IoU of 0.574 and 0.511, respectively. This improvement in 3D-NVS is due to a better selection of the second view, given the first input image. On the other hand, both Pix2Vox and 3D-R2N2 chooses a fixed set of views for two-view reconstruction -- thus not learning subsequent views. An illustration of reconstructed shapes is provided in Figure \ref{fig:sample_ouputs} where we also show reconstructed shapes when we choose $i^{th}$ view for $i>1$. We can observe that for 3D-NVS, we obtain a good reconstruction (IoU $\approx$ 0.80) while the shape degrades as we choose less probable next views.

Next, we use the texture-rich, limited-elevation rendered images used in Pix2Vox, 3D-R2N2, and Yang et al. and trained our model on that rendered images. We obtain a mean IoU of 0.651, which is better than two-view reconstruction in 3D-R2N2 (0.607) and Yang et al. (0.636) and less than Pix2Vox (0.686).

In conclusion, we see that 3D-NVS outperforms all the other methods in both reconstructions using images captured from spread-out viewpoints and the ability to generalize to different input image settings. When moving from a texture-rich, limited-elevation data to a spread-out view, the performance of Pix2Vox and 3D-R2N2 drops by 11\% and 9\%, respectively. In contrast, this decrease in performance is only 5\% in 3D-NVS. We could not compare results in Yang et al. on our rendered images due to the unavailability of open-sourced reproducible code. Most of the other state of the art techniques use range cameras with depth map for reconstruction or view selection, and comparison with those methods is not relevant since 2D images contain much less information than RGBD images.
We also compare our view selection method with commonly used baselines in the literature, namely 3D-NVS-R, where we choose next views randomly, and 3D-NVS-F, which is choosing views so as to maximize the spatial distance between camera positions. This choice of view selection is a heuristic widely used in next view prediction \cite{shapenet,yang2018active,delmerico2018comparison, hashim2019next, ly2019autonomous}. A detailed analysis of the setting and results is included in the Ablation Study (see Section \ref{sec:ablation}). Moreover, we are able to do this in real-time, unlike traditional methods in next view prediction, which use range sensors and optimize for surface coverage or information gain and take tens of seconds. This high speed of our approach is possible as it only requires a forward pass through the NVS-net and can also be highly parallelized using a GPU.
\subsubsection{Next-View Prediction}
\begin{figure}
\begin{center}
\includegraphics[scale=0.37]{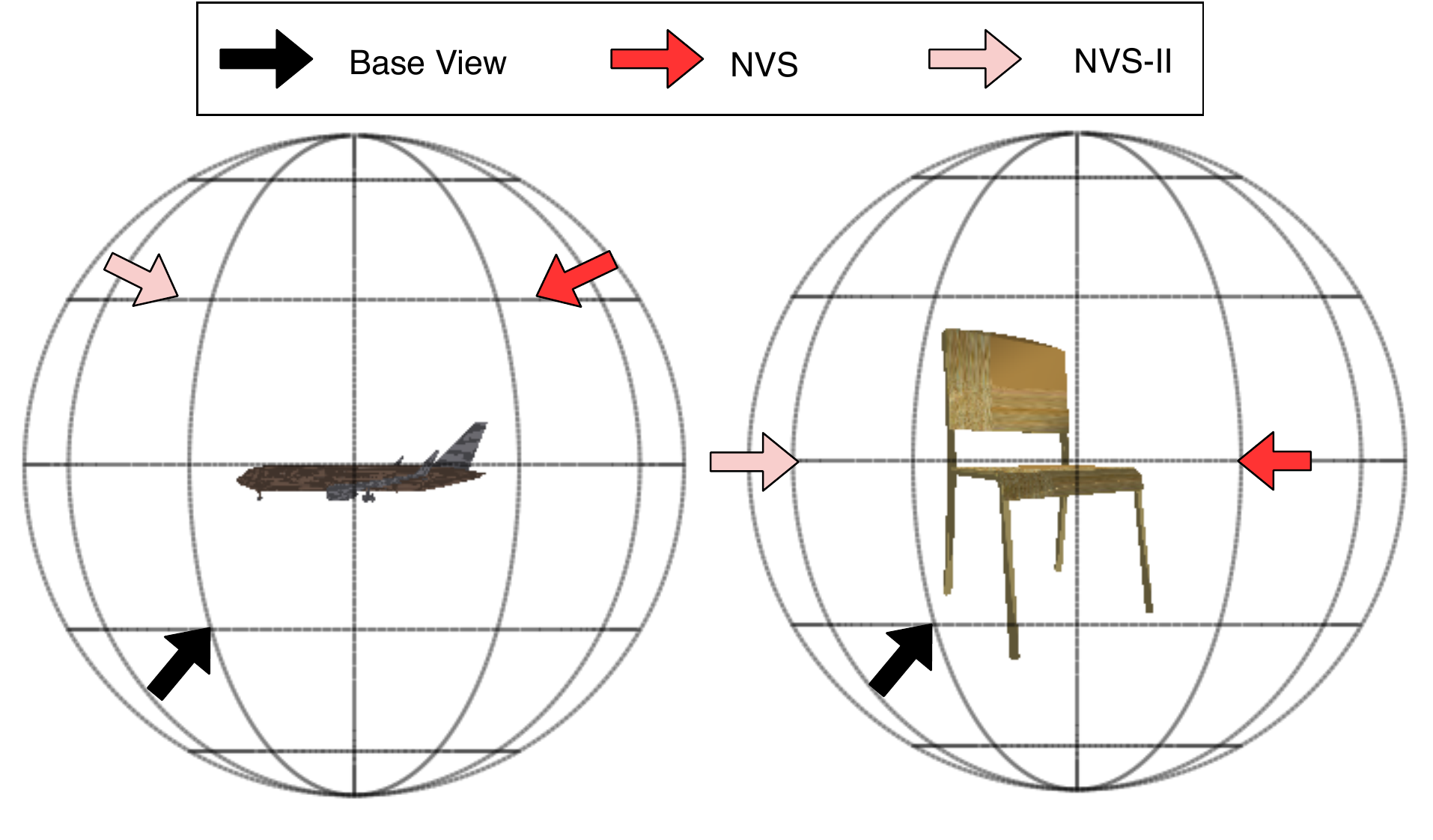}
\end{center}
   \caption{Example showing locations of next view selected by the proposed method for two object classes aeroplane and chair. For aeroplane class, both the best and second-best views are majorly above the equatorial position. For chair, it is indeed the equatorial position, which concurs with surface methods of view selection.}
\label{fig:next-view-pattern}
\end{figure}

One of the advantages of our 3D-NVS over other previous methods is that we do not require ground-truth for next view selection despite it being posed as a classification problem. Consequently, our method learns this selection on the fly using supervision from 3D shape. It is expected that for some classes of objects, viewing from a particular position is generally more informative than some other viewpoints. For example, consider the class aeroplane. If we view an aeroplane from the equatorial plane, a significantly less area is captured in the image. Thus, it is less likely that the chosen next view will lie on the equatorial plane. Instead, viewing from an elevation will be more informative. Now consider the class chair. Intuitively, the maximum area is exposed when it is viewed from an equatorial plane. 

3D-NVS can learn this class-specific next view selection without explicit characterization. We illustrate this effect with two example classes, as shown in Figure \ref{fig:next-view-pattern}. For both the classes, we choose a fixed position for the base image. Next, we observe the best and second-best selection of NVS-Net in the test dataset, which we denote as 3D-NVS and 3D-NVS-II, respectively. As expected, we observe that NVS is based on the 3D shape of the object for both aeroplanes and chairs. These observations are in conjunction with works that use range images and choose views to maximize the surface area scanned by the sensor reinforcing the effectiveness of 3D-NVS.
\vspace{-2mm}
\subsubsection{Real Images}
We also test our model on real images from the Pix3D dataset. We generate a mapping of the next views as described previously in section \ref{sec:dataset}. We encounter some missing images as Pix3D consists of common items, and the images are taken from typical standing positions. Thus, we choose the next view based on the best NVS-Net output to map the next image. Denoting $Y$ as the output of NVS-Net and $A$ as a boolean vector corresponding to available viewpoints, the next view selection is 
\begin{equation}
   \argmax_i A_i Y_i
\end{equation}
We do not separately train the network again on real images unlike in Pix3D \cite{pix3d}, 3D-R2N2 \cite{3dr2n2} and Pix2Vox \cite{pix2vox}. 3D-NVS achieves an IoU of 0.105, while randomized view selection (NVS-R) achieves an IoU of 0.08. A similar performance is obtained in Pix2Vox when a model trained on synthetic images is used. Thus, we see that using NVS-Net for next view prediction improves the performance on real images even without explicit training for such images. Therefore, 3D-NVS learns to better generalized on unseen data and seems to be better at scene understanding.

\subsection{Ablation Study}
\label{sec:ablation}
In this section, we discuss the performance comparison when certain parts of 3D-NVS are replaced with other view selection methods to show our network's effectiveness. 
\subsubsection{Individual Sub-network Performance}
Our proposed 3D-NVS selects the next view based on features extracted from the base image. In contrast, some earlier work uses camera position as a parameter to select the next view. We show that if NVS-net is removed and replaced with these heuristics, the performance decreases.

We use two methods for view selection: random next view (3D-NVS-R) and farthest next view (3D-NVS-F); both are widely used baseline in next view prediction \cite{shapenet,yang2018active,delmerico2018comparison, hashim2019next, ly2019autonomous}. In the former, given any input image, we choose the next image randomly from all the available views. On the other hand, for the farthest next view, we incorporate camera position knowledge and find out the next view which is spatially the farthest. In both cases, we observe that the mean reconstruction IoU is significantly less than the IoU obtained when using NVS-net output to select the next views. The results are presented in Table \ref{tab:result}.

Another important feature of 3D-NVS is the Context-Aware Fusion, followed by Refiner. We again observe that if we replace these sub-parts and replace them with a simple average of coarse volumes, the performance decreases substantially. These findings are in coherence with the findings of Pix2Vox \cite{pix2vox}, and hence, we omit a numerical comparison.
\vspace{-1mm}
\subsubsection{View selection orders}
\begin{figure}
\begin{center}
\includegraphics[scale=0.42]{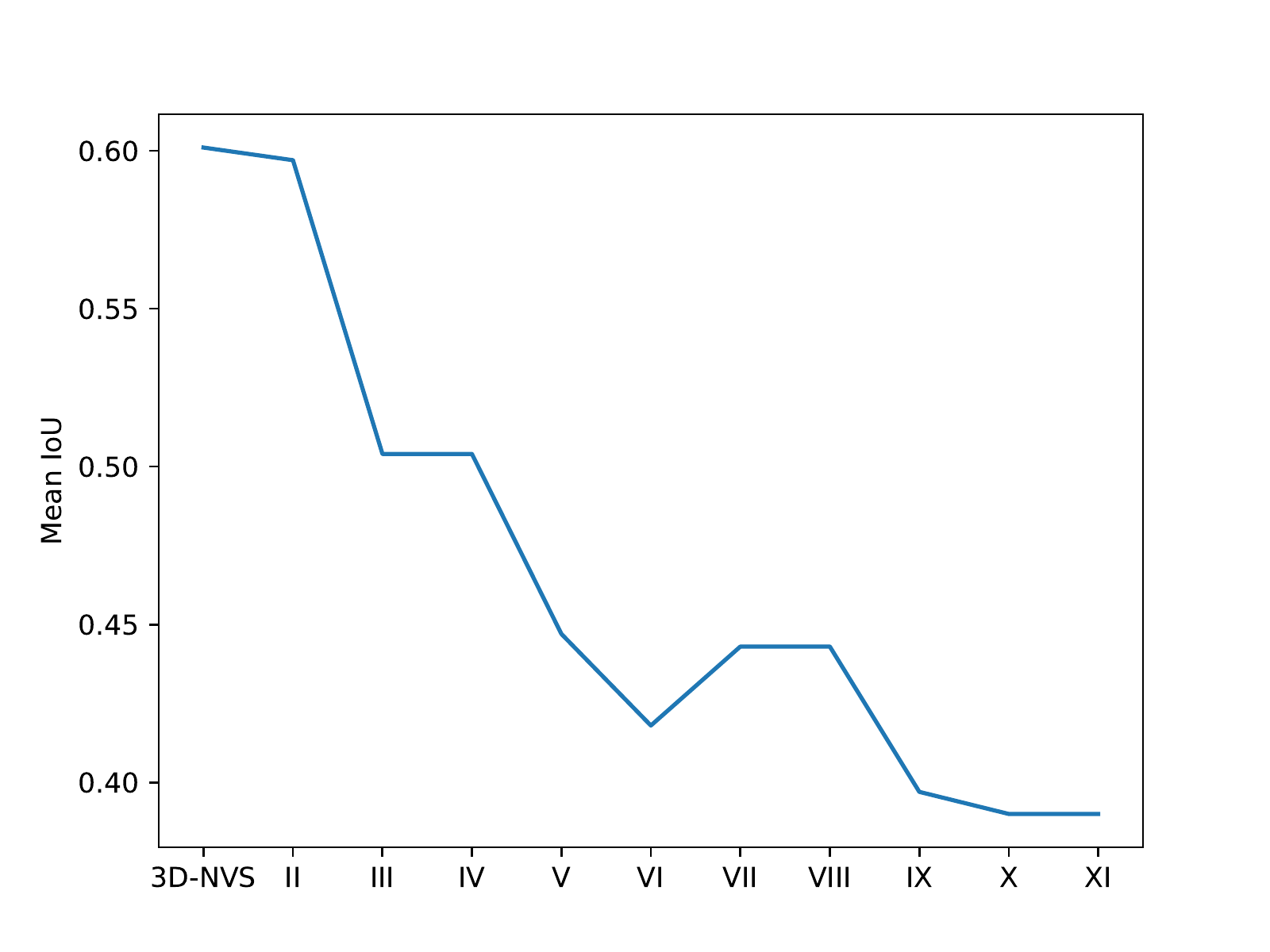}
\end{center}
   \caption{A plot of how the reconstruction performance varies with sub-optimal view selections from 3D-NVS. Here, 3D-NVS-II denotes second best view selected, 3D-NVS-III denotes the third best view and so on. We can observe that the mean IoU monotonically decreases which reinforces that 3D-NVS successfully learns to rank the views in a decreasing order of goodness.}
\label{fig:ablation}
\end{figure}

During testing, the NVS-Net gives a softMax output, which is the probability distribution over the next views. The next view with the highest probability is selected as the next best view. Now, to investigate if the NVS-Net indeed outputs the best possible view selection, we modify the network to choose the $i^{th}$-best view ($i>1$) and compare the results with NVS-Net. All the other parts of the network are kept the same.
We observe a decrease in the mean IoU of the reconstruction as we increase $i$. The results are plotted in Figure \ref{fig:ablation} and an example of reconstructed shapes is given in Figure \ref{fig:sample_ouputs}. This decrease in performance when we deter from the selection of the NVS-Net implies that NVS-Net is indeed effective in selecting the next view.

\section{Conclusion}
In this paper, we present how the task of Next Best View selection can be posed as a classification problem and learned in a data-driven manner without the knowledge of underlying ground truth labels.
We do this intending to obtain the best possible 3D reconstruction from the acquired views and use this reconstruction process to obtain indirect supervision to train the proposed model.
The proposed 3D-NVS method's performance is evaluated on both synthetic and real data to demonstrate how it obtains better reconstruction quality than the existing state of the art 3D reconstruction techniques by effectively selecting the best view.
Additionally, we present an analysis of how the selected next view locations are distributed intuitively depending on the object category, which concurs with previous techniques that use range sensors.
Going forward, we aim to extend this to successive N-view next view selection.

{\small
\bibliographystyle{ieee_fullname}
\bibliography{refs}
}

\end{document}